\documentclass[10pt]{article} 
\usepackage[preprint]{tmlr}

\usepackage{amsmath,amsfonts,bm}

\def\eqref#1{equation~\ref{#1}}

\def\1{\bm{1}}

\DeclareMathAlphabet{\mathsfit}{\encodingdefault}{\sfdefault}{m}{sl}
\SetMathAlphabet{\mathsfit}{bold}{\encodingdefault}{\sfdefault}{bx}{n}

\usepackage[hidelinks]{hyperref}
\usepackage{url}

\usepackage{algorithm}
\usepackage{algorithmic}
\usepackage{tikz}
\usetikzlibrary{arrows.meta, positioning, shapes, calc}

\title{Towards Emotionally Intelligent and Responsible \\ Reinforcement Learning}

\author{\name Garapati Keerthana \email p20240505@hyderabad.bits-pilani.ac.in \\
      \addr Department of Computer Science\\
      BITS Pilani Hyderabad
      \\
      \\
      \name Manik Gupta \email manik@hyderabad.bits-pilani.ac.in \\
      \addr Department of Computer Science\\
      BITS Pilani Hyderabad}

\defcitealias{keerthana2025dense}{DENSE}
\defcitealias{keerthana2025cli}{Cli-RAG}
\defcitealias{keerthana2025trimedprompt}{TriMedPrompt}

\begin{document}

\maketitle

\begin{abstract}
Personalized decision systems in healthcare and behavioral support often rely on static rule-based or engagement-maximizing heuristics that overlook users' emotional context and ethical constraints. Such approaches risk recommending insensitive or unsafe interventions, especially in domains involving serious mental illness, substance use disorders, or depression. To address this limitation, we propose a \textit{Responsible Reinforcement Learning} (RRL) framework that integrates emotional and contextual understanding with ethical considerations into the sequential decision-making process. RRL formulates personalization as a Constrained Markov Decision Process (CMDP), where the agent optimizes engagement and adherence while ensuring emotional alignment and ethical safety. We introduce a multi-objective reward function that explicitly balances short-term behavioral engagement with long-term user well-being, and define an emotion-informed state representation that captures fluctuations in emotional readiness, affect, and risk. The proposed architecture can be instantiated with any RL algorithm (e.g., DQN, PPO) augmented with safety constraints or Lagrangian regularization. Conceptually, this framework operationalizes empathy and responsibility within machine learning policy optimization, bridging safe RL, affective computing and responsible AI. We discuss the implications of this approach for human-centric domains such as behavioral health, education, and digital therapeutics, and outline simulation-based validation paths for future empirical work. This paper aims to initiate a methodological conversation about ethically aligned reinforcement learning for emotionally aware and trustworthy personalization systems.
\end{abstract}

\section{Introduction}

Personalization is a defining feature of intelligent digital systems, from recommender engines to behavioral health platforms \cite{li2010contextual,hornstein2023personalization,bond2023digital}. However, many current systems rely on static heuristics or engagement-optimized models that treat users as homogeneous agents \cite{stray2024building}. In high-stakes contexts—such as behavioral health, addiction recovery, or emotional support—this approach can result in ethically inappropriate or emotionally misaligned recommendations \cite{valentine2023recommender}. For individuals experiencing serious mental illness or psychological distress, a decision that maximizes short-term engagement may inadvertently cause harm, exacerbate vulnerability, or erode trust \cite{grabb2024risks}. This tension highlights a pressing need for personalization frameworks that are not only adaptive and data-driven but also responsible, empathetic, and contextually aware.

Reinforcement Learning (RL) provides a natural paradigm for adaptive personalization by modeling sequential decision-making and long-term reward optimization \cite{den2020reinforcement}. Yet conventional RL formulations typically assume that all rewards are desirable, measurable, and ethically neutral. When deployed in human-centric domains, this assumption fails: optimizing engagement alone neglects nuanced human factors such as emotional readiness, safety, and psychological well-being. Moreover, standard RL agents lack a mechanism to interpret or act upon users' affective cues—information critical for building trust and ensuring appropriate interventions.

To bridge this gap, we propose the concept of \textit{Responsible Reinforcement Learning} (RRL), a principled extension of RL that incorporates emotional context and ethical constraints directly into policy optimization. The central idea is that a responsible agent should learn not only \emph{what} action maximizes long-term utility, but also \emph{when} and \emph{for whom} an action is emotionally or ethically appropriate. Our framework augments traditional RL with (1) an emotion-informed state representation that encodes affective and contextual signals, and (2) a multi-objective reward function that explicitly balances engagement, empathy, and safety. Additionally, we introduce constraint-aware optimization to enforce hard safety limits, ensuring that the agent’s actions remain within ethical boundaries.

\begin{figure}[ht]
\centering
\begin{tikzpicture}[font=\small, 
    node distance=1cm and 1cm,
    every node/.style={align=center},
    venn circle/.style={circle, minimum width=4cm, fill opacity=0.25, text opacity=1, draw, thick}
]

\node[venn circle, fill=blue!40, label={[align=center]above:\textbf{Safe Reinforcement Learning}\\(Safety, Risk-Aware Optimization)}] (A) at (0,0) {};
\node[venn circle, fill=green!40, label={[align=center]above right:\textbf{Affective Computing}\\(Emotion, Empathy, Context)}] (B) at (2.5,0) {};
\node[venn circle, fill=orange!40, label={[align=center]below:\textbf{Responsible AI}\\(Fairness, Transparency, Ethics)}] (C) at (1.25,-2.2) {};

\begin{scope}
    \clip (A) circle (2cm);
    \clip (B) circle (2cm);
    \clip (C) circle (2cm);
    \fill[magenta!40, opacity=0.5] (1.25,-0.8) circle (1cm);
\end{scope}

\coordinate (center) at (1.25,-0.8);
\coordinate (exit) at (6,-0.8); 

\node[draw=black, thick, fill=white, rounded corners, inner sep=6pt, text width=7.5cm, align=center, font=\small\bfseries] (rrl) at (10,-0.8)
{Responsible Reinforcement Learning (RRL)\\Integrates Empathy, Safety and Ethics};

\draw[dotted, very thick, black!70!black, -{Latex[length=3mm]}] (center) -- (exit) -- (rrl.west);

\end{tikzpicture}

\caption{Conceptual synthesis of \textbf{Safe Reinforcement Learning}, \textbf{Affective Computing}, and \textbf{Responsible AI}. Their intersection (highlighted region) gives rise to the proposed \textbf{Responsible Reinforcement Learning (RRL)} framework, which operationalizes responsibility as a measurable component of adaptive decision-making.}
\label{fig:RRL_concept}
\end{figure}
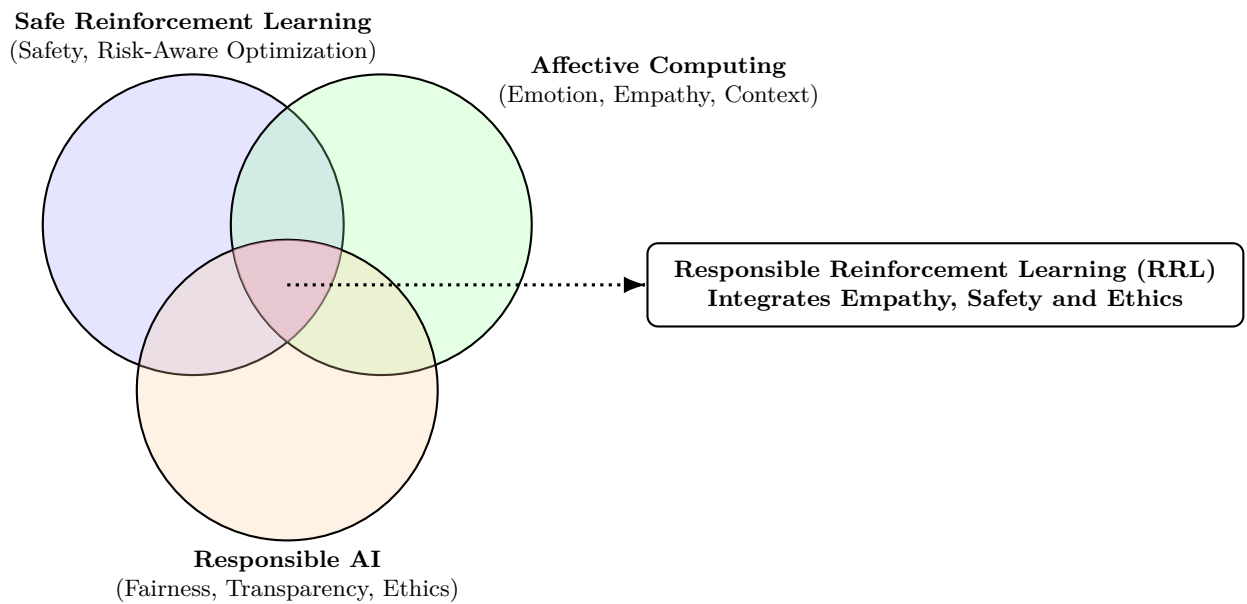

This formulation operationalizes responsibility as a measurable component of learning—moving beyond abstract AI ethics principles toward concrete algorithmic mechanisms. Conceptually, the proposed framework unifies insights from three active research areas as shown in \ref{fig:RRL_concept} \textbf{Safe Reinforcement Learning}, which aims to prevent harmful exploration; \textbf{Affective Computing}, which models emotional states in human-AI interaction; and \textbf{Responsible AI}, which emphasizes transparency, fairness, and harm reduction. By integrating these perspectives, RRL provides a foundation for developing AI systems capable of empathetic, trustworthy, and ethically aligned personalization.

Our contributions are threefold:
\begin{enumerate}
    \item We introduce a conceptual and mathematical framework for \textbf{Responsible Reinforcement Learning (RRL)}, integrating emotion-aware representations and ethical constraints into policy optimization.
    \item We propose a \textbf{multi-objective reward formulation} that captures the trade-off between engagement, empathy, and safety, enabling adaptive yet responsible personalization.
    \item We outline a simulation-based experimental design for evaluating RRL in human-centric settings without access to sensitive real-world data, thereby facilitating reproducibility and ethical research practices.
\end{enumerate}

Through this work, we aim to initiate a dialogue between technical RL research and the emerging field of responsible AI for human well-being. The RRL paradigm lays the groundwork for a new generation of agents that are not merely intelligent, but also emotionally intelligent, ethically constrained, and socially aware.

\section{Related Work}
\label{sec:related}

Our proposed Responsible Reinforcement Learning (RRL) framework draws on three intersecting literatures: (1) safe/constrained reinforcement learning, (2) affective computing and emotional state modeling in human-AI interaction, and (3) responsible AI / governance frameworks for high-stakes deployments. Below we summarize representative work in each area and highlight the gaps that RRL aims to address.

\subsection{Safe and Constrained Reinforcement Learning}
Safety in reinforcement learning has been an active research area for over a decade. Surveys and reviews categorize safe-RL methods into approaches that modify the optimality criterion, shape exploration, or impose runtime safety mechanisms \cite{garcia2015comprehensive,zhao2023statewise,wachi2024survey}. Constrained Markov decision processes (CMDPs) provide a formalism for encoding safety constraints, and policy-search methods that respect constraints have been developed with theoretical guarantees. A prominent algorithm in this space is Constrained Policy Optimization (CPO), which enforces near-constraint satisfaction during policy updates and enables learning neural-network policies under safety limits \cite{achiam2017constrained}. More recent work extends these ideas with Lagrangian relaxations, shielded execution, and human-in-the-loop safety monitors; however, most safe-RL research focuses on physical or robotic safety (collision avoidance, energy limits) rather than the socio-emotional safety concerns relevant to human-centered domains \cite{chow2018lyapunovSafeRL}.

\subsection{Reinforcement Learning in Human-Centric and Healthcare Domains}
Reinforcement learning has been explored for healthcare decision problems including dynamic treatment regimes, scheduling, and clinical decision support. These surveys demonstrate both the promise and the unique challenges of applying RL in clinical contexts: reward specification is difficult, feedback is delayed and noisy, safety and interpretability are critical, and real-world evaluation is constrained by ethical/regulatory concerns \cite{yu2021reinforcement}. While several demonstration studies show RL can learn useful policies in simulated or retrospective settings, there remains a gap in methods that explicitly integrate affective or emotional signals and that operationalize emotion and ethics as part of the optimization objective.

Recent advances in clinical AI highlight the value of modeling temporally evolving user or patient states and integrating heterogeneous contextual signals across interactions. Additionally, unified prompting approaches ensure that synthesized clinical notes maintain content- and layout-consistency, supporting structured and reliable representations \citepalias{keerthana2025trimedprompt}. Collectively, these works demonstrate the utility of rich, temporally structured, and context-sensitive modeling for personalized decision support.

Such insights motivate the design of reinforcement learning methods that could, in principle, incorporate analogous representations of user state, engagement history, and emotional context. By leveraging temporally and contextually informed representations, RL in human-centric domains may better reason about both intervention effectiveness and emotional alignment, capturing trade-offs between engagement, empathy, and ethical considerations, even when direct behavioral data is limited.

\subsection{Affective Computing and Emotion-Aware Systems}
Affective computing seeks to detect, model, and respond to human emotions; it provides the perceptual and representational tools necessary to make personalization emotionally aware \cite{picard1997affective}. Work on relational agents and emotionally adaptive conversational systems has shown that systems capable of exhibiting empathy or emotional sensitivity (e.g., virtual coaches and relational agents) can improve engagement and user satisfaction in health-related tasks \cite{bickmore2005relational,bickmore2010maintaining, nelekar2022effectiveness}. More recent surveys of affective computing in digital mental health confirm growing interest in emotion-aware interventions, but they also report limited evidence on long-term clinical outcomes and emphasize ethical concerns such as privacy and bias \cite{affective_survey_2022,affective_review_2025}. Importantly, most affective-computing systems operate as perception-then-response pipelines and are not tightly integrated with sequential decision frameworks that optimize long-term outcomes under safety constraints.

\subsection{Responsible AI and Risk Management}
Responsible AI frameworks (e.g., the NIST AI Risk Management Framework (RMF) \cite{tabassi2023aiRMF} and the IndiaAI Governance Framework \cite{indiaAI2025governance}); various academic and policy guidelines) establish principles and practices—transparency, fairness, accountability, risk management—that are critical for deploying AI in sensitive domains. These frameworks emphasize the need for risk-based governance, documentation, and monitoring, yet they are largely agnostic to algorithmic design. Translating principles into algorithmic mechanisms (for example, embedding fairness or empathy into objective functions or enforcing safety constraints during learning) remains an open area of research.

\subsection{Gaps and Positioning}
The literature provides strong foundations: (a) formal methods for constrained RL and safety-aware policy learning \cite{garcia2015comprehensive,achiam2017constrained}, (b) perceptual models and evidence that emotional sensitivity affects engagement \cite{picard1997affective,bickmore2005relational}, and (c) high-level governance frameworks for trustworthy AI \cite{tabassi2023aiRMF}. However, three gaps persist:

\begin{enumerate}
  \item \textbf{Operationalizing emotion within RL:} Existing safe-RL work rarely incorporates affective state as a first-class input to the decision process; affective computing systems are typically used for single-turn adaptation, rather than long-horizon policy optimization.
  \item \textbf{Multi-objective optimization for empathy and safety:} There is limited work that treats empathy, emotional alignment, and human safety as explicit objectives or  constraints alongside conventional engagement/reward metrics.
  \item \textbf{Practical evaluation without sensitive data:} Ethical and regulatory barriers restrict access to real behavioral-health datasets; there is a need for reproducible simulation protocols and evaluation metrics that reflect empathy and safety trade-offs.
\end{enumerate}

Our RRL framework addresses these gaps by (1) defining an \emph{emotion-informed state representation} suitable for RL, (2) proposing a \emph{multi-objective reward and constraint} formulation that places safety and emotional alignment on par with engagement, and (3) providing a simulation-based experimental design for researchers to evaluate policies reproducibly without relying on private clinical data. In the next section we formalize the RRL model and describe algorithmic instantiations that leverage constrained optimization and perception modules for emotion inference.

\section{Details of RRL Framework}
\label{sec:method}

This section formalizes the proposed \textit{Responsible Reinforcement Learning} (RRL) framework and explains how it extends standard reinforcement learning to incorporate ethical and emotional considerations. We begin by introducing the Constrained Markov Decision Process (CMDP) formulation, then describe the emotion-informed state representation, multi-objective reward design, and constraint-aware policy optimization. Finally, we discuss the rationale for adopting a CMDP and present a concise algorithmic summary.

\subsection{From Reinforcement Learning to Responsible Decision-Making}

Reinforcement Learning (RL) provides a mathematical foundation for sequential decision-making under uncertainty \cite{yu2021reinforcement, liu2024review}. An RL agent interacts with an environment at discrete time steps, observing a state $s_t$, taking an action $a_t$, and receiving a scalar reward $R(s_t,a_t)$. The objective is to learn a policy $\pi(a|s)$ that maximizes the expected discounted cumulative reward:
\begin{equation}
J(\pi) = \mathbb{E}_\pi \left[\sum_{t=0}^{T-1} \gamma^t R(s_t,a_t)\right],
\end{equation}
where $\gamma \in (0,1)$ is a discount factor controlling the trade-off between short- and long-term outcomes.

\begin{figure}[ht]
    \centering
    \begin{tikzpicture}[
        node distance=1.6cm and 2cm,
        every node/.style={align=center, font=\small},
        user/.style={rectangle, rounded corners, draw=blue!60, thick, fill=blue!10, text width=3.2cm, minimum height=1cm},
        emo/.style={rectangle, rounded corners, draw=purple!60, thick, fill=purple!10, text width=3.5cm, minimum height=1cm},
        agent/.style={rectangle, rounded corners, draw=orange!80, thick, fill=orange!10, text width=3.5cm, minimum height=1cm},
        act/.style={rectangle, rounded corners, draw=green!60!black, thick, fill=green!10, text width=3cm, minimum height=1cm},
        arrow/.style={-Stealth, thick}
    ]
        \node[user] (user) {User Profile \\ (demographics, diagnosis)};
        \node[emo, below=1.5cm of user] (emotion) {Emotional Readiness Encoder \\ Affective state $e_t$};
        \node[user, left=of emotion] (history) {Behavioral History \\ (engagement \& interaction data)};
        \node[agent, below=1.5cm of emotion] (agent) {Responsible RL Agent \\ $\pi(a|s)$ with ethical constraints};
        \node[act, right=1.5cm of agent] (action) {Personalized \\ \textbf{Ethical Action}};

        \draw[arrow] (user.south) -- (emotion.north);
        \draw[arrow] (history.east)  -- (emotion.west);
        \draw[arrow] (emotion.south) -- (agent.north);
        \draw[arrow] (agent.east) -- (action.west);
    \end{tikzpicture}
    \caption{Conceptual overview of the proposed Responsible Reinforcement Learning (RRL) framework. The agent integrates emotional and contextual information to recommend safe, personalized interventions.}
    \label{fig:rrl_framework}
\end{figure}
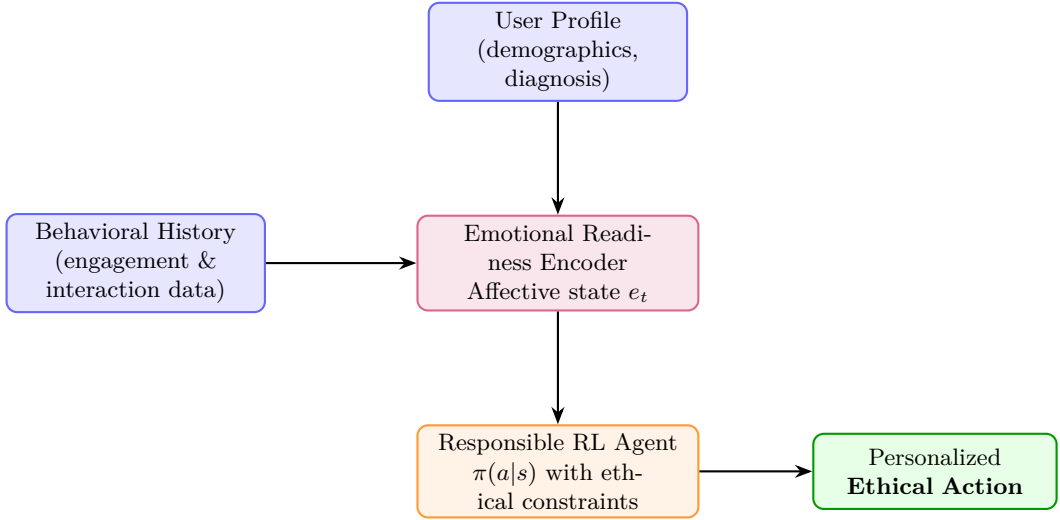

However, in sensitive applications such as behavioral health, maximizing engagement alone can lead to emotionally inappropriate or ethically unsafe recommendations.  framework as shown in Figure~\ref{fig:rrl_framework} extends standard RL by introducing explicit constraints that encode safety, empathy, and fairness, ensuring that optimization respects human well-being.

\subsection{Constrained Markov Decision Process Formulation}

We model the environment as a Constrained Markov Decision Process (CMDP) \cite{achiam2017constrained, garcia2015comprehensive}, represented by the tuple $(\mathcal{S}, \mathcal{A}, P, R, C, \gamma)$ where:
\begin{itemize}
  \item $\mathcal{S}$ denotes the state space.
  \item $\mathcal{A}$ represents the action space.
  \item $P(s_{t+1} \mid s_t, a_t)$ defines transition probabilities.
  \item $R(s_t, a_t)$ is the primary reward function capturing engagement or adherence outcomes.
  \item $C(s_t, a_t)$ is a cost function capturing ethical or emotional violations.
  \item $\gamma$ is the discount factor.
\end{itemize}

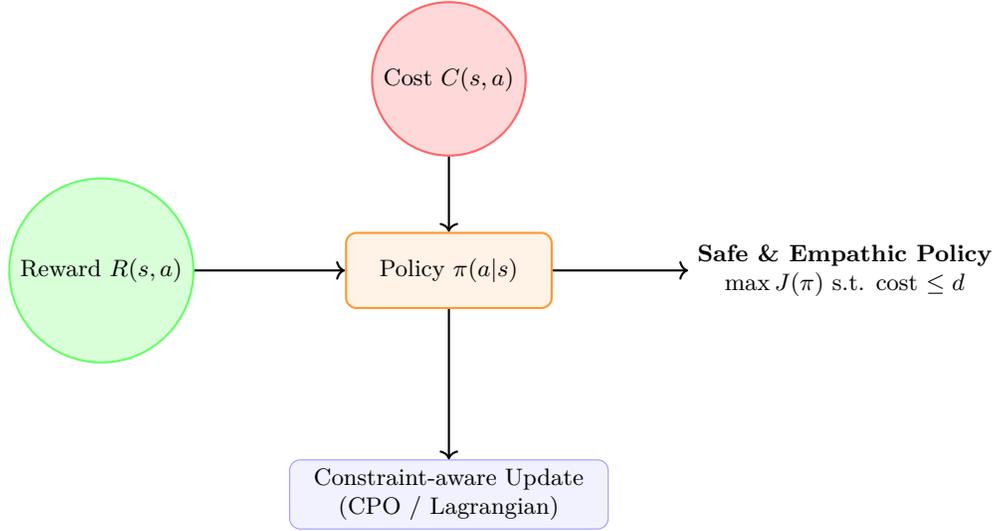
\begin{figure}[t]
    \centering
    \begin{tikzpicture}[
        every node/.style={align=center, font=\small},
        reward/.style={circle, draw=green!60, fill=green!15, thick, minimum size=1.5cm},
        cost/.style={circle, draw=red!60, fill=red!15, thick, minimum size=1.5cm},
        policy/.style={rectangle, rounded corners, draw=orange!80, fill=orange!10, thick, text width=2.5cm, minimum height=1cm},
        arrow/.style={->, thick}
    ]
        \node[policy] (policy) {Policy $\pi(a|s)$};
        \node[reward, left=2cm of policy] (reward) {Reward $R(s,a)$};
        \node[cost, above=1cm of policy] (cost) {Cost $C(s,a)$};
        \node[rectangle, draw=blue!40, fill=blue!5, rounded corners, text width=4cm, below=2cm of policy] (update) {Constraint-aware Update \\ (CPO / Lagrangian)};
        \node[right=1.8cm of policy] (goal) {\textbf{Safe \& Empathic Policy} \\ $\max J(\pi)$ s.t. cost $\le d$};

        \draw[arrow] (reward.east) -- (policy.west);
        \draw[arrow] (cost.south) -- (policy.north);
        \draw[arrow] (policy.south) -- (update.north);
        \draw[arrow] (policy.east) -- (goal.west);
    \end{tikzpicture}
    \caption{CMDP-based optimization within RRL. The agent learns to balance reward (engagement) and cost (emotional or ethical risk) through constraint-aware updates.}
    \label{fig:cmdp}
\end{figure}

The goal is to learn a policy $\pi(a|s)$ that maximizes expected cumulative reward while maintaining the expected cumulative cost below a threshold $d$:
\begin{equation}
\max_{\pi} \; \mathbb{E}_\pi \left[\sum_{t=0}^{T-1} \gamma^t R(s_t,a_t)\right]
\quad \text{s.t.} \quad
\mathbb{E}_\pi \left[\sum_{t=0}^{T-1} \gamma^t C(s_t,a_t)\right] \le d.
\end{equation}

where $d$ is a predefined \textit{ethical or emotional risk threshold} dependant on the domain and usecase. It specifies the maximum acceptable level of cumulative cost that the agent may incur across its trajectory. In behavioral health personalization, $d$ can represent the tolerance for emotionally inappropriate recommendations, user distress, or clinically unsafe actions. Adjusting $d$ effectively tunes the balance between exploratory engagement and user safety. This formulation as shown in Figure ~\ref{fig:cmdp} explicitly enforces ethical boundaries during learning, providing a mathematically grounded mechanism for responsible personalization.

\subsection{Worked Example and Formal Guarantee Considerations}
To illustrate the proposed emotionally-informed CMDP formulation in concrete terms, consider a minimal toy environment:
\[
\mathcal{M} = (\mathcal S, \mathcal A, T, R, C),
\]
with \(\mathcal S = \{s_{\rm neutral}, s_{\rm emotional}\}\), \(\mathcal A = \{a_{\rm engage}, a_{\rm disengage}\}\), transition kernel \(T\), reward \(R\), cost \(C\) defined as:
\[
R(s_{\rm neutral},a_{\rm engage}) = r_0,\;\; C(s_{\rm emotional},a_{\rm engage}) = c_1,
\]
and analogous definitions for other combinations.  
One may define the Lagrangian objective
\[
\max_\pi \; \mathbb E_{\pi} \left[ \sum_{t=0}^\infty \gamma^t R(s_t,a_t) \right]
\quad\text{s.t.}\quad
\mathbb E_{\pi} \left[ \sum_{t=0}^\infty \gamma^t C(s_t,a_t) \right] \le d.
\]
Although this model is simplified, it highlights how the emotional-state cost can be mapped into the CMDP cost term.  
We note that current literature on constrained RL provides theoretical convergence \cite{garcia2015comprehensive} under certain assumptions (e.g., strong convexity, bounded cost functions), yet stopping-time or safe-exploration guarantees remain challenging \cite{krasowski2023provably}  
Future work must extend such guarantees to our emotionally-informed state space and composite reward structure.

\noindent\textbf{Takeaway:} A formal guarantee of the form “for any \(\varepsilon>0\), the policy \(\pi\) will satisfy the cost bound with probability at least \(1-\delta\)" would significantly strengthen the contribution.

\subsection{Emotion-Informed State Representation}

Standard RL frameworks rely on observable behavioral or demographic features, which are often insufficient to capture emotional readiness or sensitivity. In RRL, the state is expanded to include affective information:
\begin{equation}
s_t = [u_t, h_t, e_t],
\end{equation}
where $u_t$ denotes static or slowly evolving user attributes (e.g., diagnosis, demographics), $h_t$ captures behavioral engagement history, and $e_t$ is an \textit{emotion-informed embedding} derived from contextual cues such as text sentiment, voice tone, or self-reported readiness. This embedding is obtained through a separate affective perception module, possibly based on pretrained emotion classifiers or transformer-based encoders \cite{picard1997affective, schuller2024multimodal, schlicher2025emotionally}. In addition, relevant contextual information can be retrieved from prior interactions or external knowledge sources using retrieval-augmented transformer models \citepalias{keerthana2025cli}, enabling the embedding to incorporate temporally and contextually aligned textual cues for more accurate emotion representation.
 
By incorporating $e_t$, the policy gains the ability to anticipate the affective impact of its actions and adjust decisions to preserve empathy and safety. The emotional representation also facilitates interpretability, allowing practitioners to understand how affective context influences policy behavior.

\subsection{Multi-Objective Reward Design}

Because emotional appropriateness and engagement objectives may not always align, RRL employs a composite reward formulation:
\begin{equation}
R_{\text{composite}}(s_t,a_t) = 
w_{\text{eng}} \, r_{\text{eng}}(s_t,a_t)
+ w_{\text{emo}} \, r_{\text{emo}}(s_t,a_t)
- w_{\text{safety}} \, \mathbf{1}\{\text{safety\_violation}(s_t,a_t)\},
\end{equation}
where $r_{\text{eng}}$ represents engagement-based reward, $r_{\text{emo}}$ measures emotional alignment between user state and action, and the final term penalizes ethical or safety violations. The coefficients $w_{\text{eng}}, w_{\text{emo}},$ and $w_{\text{safety}}$ determine the trade-offs between engagement and responsibility.

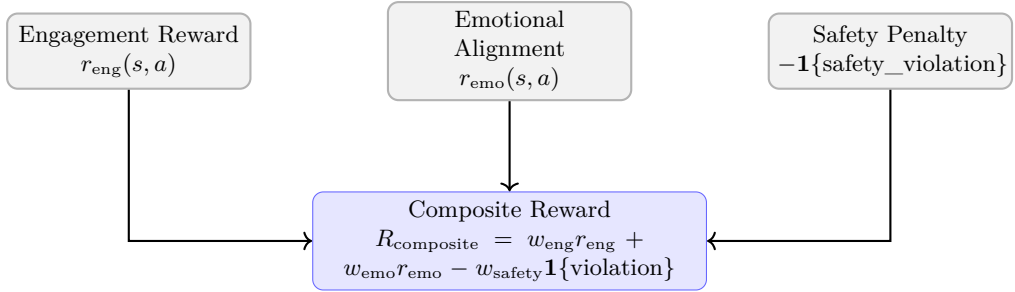
\begin{figure}[t]
    \centering
    \begin{tikzpicture}[
        every node/.style={align=center, font=\small},
        comp/.style={rectangle, rounded corners, draw=gray!60, thick, text width=3cm, fill=gray!10, minimum height=1cm},
        arrow/.style={->, thick}
    ]
        \node[comp] (eng) {Engagement Reward \\ $r_{\text{eng}}(s,a)$};
        \node[comp, right=of eng, xshift=0.8cm] (emo) {Emotional Alignment \\ $r_{\text{emo}}(s,a)$};
        \node[comp, right=of emo, xshift=0.8cm] (safe) {Safety Penalty \\ $-\mathbf{1}\{\text{safety\_violation}\}$};
        \node[rectangle, draw=blue!60, fill=blue!10, rounded corners, text width=5cm, below=1.2cm of emo] (compR) {Composite Reward \\ $R_\text{composite} = w_{\text{eng}}r_{\text{eng}} + w_{\text{emo}}r_{\text{emo}} - w_{\text{safety}}\mathbf{1}\{\text{violation}\}$};

        \draw[arrow] (eng.south) |- (compR.west);
        \draw[arrow] (emo.south) -- (compR.north);
        \draw[arrow] (safe.south) |- (compR.east);
    \end{tikzpicture}
    \caption{Multi-objective reward structure balancing engagement, emotional appropriateness, and safety in RRL.}
    \label{fig:reward}
\end{figure}

This reward decomposition as shown in Figure~\ref{fig:reward} enables the agent to learn policies that optimize holistic well-being rather, than single-dimensional behavioral metrics. Importantly, the model can be tuned to prioritize safety when ethical risk tolerance is low, or increase exploration when emotional uncertainty is high.

\subsection{Adaptive Weighting of Composite Rewards}
The proposed composite reward structure has three scalar weights \(w_{\rm eng}, w_{\rm emo}, w_{\rm safety}\) and rather than fixing these a priori, we suggest adopting a multi‐objective optimisation perspective. For instance, one may compute the Pareto front of policies that trade off engagement vs. emotional alignment vs. safety (Rodríguez-Chía et al., 2022). Alternatively, one could treat \(\mathbf w = (w_{\rm eng},w_{\rm emo},w_{\rm safety})\) as learnable parameters via a meta‐learning scheme:
\[
\min_{\mathbf w}\; \mathcal{L}_{\rm multi}(\pi_{\mathbf w}) 
\quad\text{s.t.}\quad \mathbb E_{\pi_{\mathbf w}}[C(s_t,a_t)]\le d.
\]
These strategies help mitigate arbitrary manual tuning and ensure alignment across objectives\cite{rodriguezchia2022multiobjective}.

\subsection{Constraint-Aware Policy Optimization}

To learn under ethical constraints, we adapt methods from safe RL such as \textit{Constrained Policy Optimization} (CPO) \cite{achiam2017constrained}. The optimization problem is cast as a Lagrangian:
\begin{equation}
\mathcal{L}(\pi,\lambda) =
\mathbb{E}_\pi \!\left[\sum_t \gamma^t R_{\text{composite}}(s_t,a_t)\right]
- \lambda \left(
\mathbb{E}_\pi \!\left[\sum_t \gamma^t C(s_t,a_t)\right] - d
\right),
\end{equation}
where $\lambda$ is a Lagrange multiplier dynamically adjusted during training. When the expected cost exceeds the allowable threshold $d$, $\lambda$ increases, tightening the constraint boundary. This approach integrates seamlessly with trust-region methods (e.g., PPO) and ensures stable convergence even under multi-objective learning.

\subsection{Why Constrained RL?}
We adopt the CMDP formulation over alternatives such as regularized or multi-objective RL for three reasons. First, CMDPs enable explicit control of ethical boundaries, guaranteeing that emotionally unsafe actions remain below a quantifiable risk threshold. Second, CMDPs align with the interpretability and accountability requirements central to healthcare and responsible AI. Finally, CMDPs provide a natural interface for integrating affective states and cost-sensitive learning with established RL algorithms, enabling scalable and transparent deployment.

\subsection{Algorithmic Summary}

\begin{center}
\rule{\linewidth}{0.4pt}
\textbf{Algorithm 1: Responsible Reinforcement Learning (RRL)}
\rule{\linewidth}{0.4pt}
\begin{algorithmic}[1]
\STATE Initialize policy $\pi(a|s;\theta)$, cost threshold $d$, and multiplier $\lambda$
\FOR{iteration $= 1$ to $N$}
    \STATE Collect trajectories $\tau = (s_t, a_t, r_t, c_t)$ using $\pi$
    \STATE Estimate discounted rewards $\hat{R}$ and costs $\hat{C}$
    \STATE Update $\pi$ using constrained gradient descent or trust-region optimization
    \STATE Adjust $\lambda \leftarrow \lambda + \eta (\hat{C} - d)$
    \STATE Update emotion encoder (if jointly trained)
\ENDFOR
\end{algorithmic}
\rule{\linewidth}{0.4pt}
\end{center}

In summary, the RRL framework extends reinforcement learning to operate in emotionally and ethically sensitive contexts, ensuring that agent decisions are not only optimal but also empathetic, interpretable, and aligned with human-centered values.

\section{Experimental Setup}
\label{sec:experiments}

Although the Responsible Reinforcement Learning (RRL) framework is designed for eventual deployment in real-world behavioral support systems, it can be evaluated and analyzed initially through controlled simulations. This section outlines a conceptual experimental design, potential metrics, and ethical considerations relevant to such evaluation.

\subsection{Simulation Environment}
In the absence of real behavioral data, a synthetic environment can be constructed to emulate key user dynamics. Each simulated user is represented as a latent state vector comprising demographic attributes, engagement history, and an emotional readiness signal. The environment transitions are governed by probabilistic rules that capture typical behavioral responses, such as reduced engagement following emotionally incongruent actions or improved adherence after contextually aligned recommendations. This approach parallels simulation methods used in prior healthcare RL research \cite{yu2021reinforcement, islam2025reinforcement}  and also draws on methods for modeling temporally structured clinical states like \citepalias{keerthana2025dense}.

Specifically, an agent observes a state $s_t = [u_t, h_t, e_t]$, selects an action $a_t$ from a discrete set of supportive interventions, and receives two feedback signals: an engagement reward $r_{\text{eng}}(s_t,a_t)$ and an emotional alignment score $r_{\text{emo}}(s_t,a_t)$. Emotional misalignment results in an increased cost signal $C(s_t,a_t)$, which is propagated through the CMDP constraint. Over multiple simulated episodes, the agent learns to balance exploration and exploitation, identifying actions that maximize cumulative reward, while minimizing emotional or ethical risk.

\subsection{Simulation Setup and Evaluation Metrics}
We propose the following simulation protocol:

\begin{itemize}
  \item \textbf{Environment:} A generative synthetic user-agent interaction model in which the user emotional state \(e_t\) transitions according to
  \[
    P(e_{t+1}\mid e_t,a_t,x_t) = 
      \begin{cases}
        p_{\rm raise}, & \text{if }a_t\text{ engages sensitively}\\
        p_{\rm lower}, & \text{if }a_t\text{ disengages or ignores}\\
      \end{cases}
  \]
  with noise parameter \(\eta\in(0,1)\). Sensor noise is modelled by \(o_t = e_t + \xi_t,\; \xi_t\sim\mathcal N(0,\sigma^2)\).
  \item \textbf{Baselines:} To assess the benefits of RRL, several baselines may be defined conceptually. The first is a \textit{static rule-based policy} that mirrors conventional heuristic systems used in many behavioral health applications. The second is a \textit{standard RL agent} (e.g., PPO or DQN) trained purely on engagement reward without safety constraints. The third baseline extends this with a simple emotional penalty term, but without formal constraint enforcement. The fourth baseline extends the second baseline without safety penalty term and finally experimenting with our emotionally informed CMDP. Comparison across these baselines would demonstrate the incremental contribution of explicit constraint modeling.
  \item \textbf{Metrics:}
  The key evaluation metrics in such a study would include: (1) total engagement return, (2) cumulative emotional alignment score, and (3) rate of safety constraint violation. Ideally, RRL should achieve comparable or higher engagement while significantly reducing ethical and emotional violations. Visualization of the reward–cost trade-off curves would illustrate the agent’s ability to operate on an efficient frontier between effectiveness and responsibility.
    \begin{itemize}
      \item \emph{Engagement rate} \(= \frac1T\sum_{t=0}^{T-1} \mathbf 1\{a_t=\text{engage}\}\).
      \item \emph{Emotional alignment} \(= \frac1T\sum_{t=0}^{T-1} \mathbb I\{\operatorname{sign}(e_t)\cdot\operatorname{sign}(f(a_t))>0\}\).
      \item \emph{Safety cost} \(= \mathbb E\left[\sum_{t=0}^{T-1} C(s_t,a_t) \right]\).
      \item \emph{Pareto efficiency index} comparing policies on the (Engagement, EmotionalAlignment, Safety) tri-objective.
    \end{itemize}
 \end{itemize}

These details ensure clarity and reproducibility, addressing a key shortcoming in prior conceptual frameworks.

\section{Discussion}

\subsection{Interpretability and Human-in-the-Loop Validation}

Beyond numerical performance, interpretability and human trust are central to behavioral applications. The CMDP structure allows introspection into which states or actions contribute most to constraint violations, providing actionable insights for designers and clinicians. Moreover, simulated policies can be audited through human-in-the-loop evaluations, where domain experts assess whether suggested interventions align with ethical and emotional norms. This methodology connects algorithmic transparency with participatory design principles in digital health \cite{bickmore2005relational, schlicher2025emotionally}.

\subsection{Ethical and Practical Considerations}
A fundamental challenge in deploying emotionally aware RL agents is ensuring that the model’s perception of emotion does not reproduce bias or stereotyping. Emotional embeddings derived from text or behavior may encode cultural or linguistic biases, potentially leading to inequitable outcomes. RRL addresses this risk partially through its explicit cost function, but practical deployment must also include continuous monitoring, fairness auditing, and optional human override mechanisms. These safeguards resonate with the goals of responsible AI governance frameworks such as the NIST AI RMF \cite{tabassi2023aiRMF}.

\subsection{Summary of Implications}
The proposed RRL framework thus offers a scalable path toward ethically aligned personalization. By incorporating emotion-informed states, multi-objective reward shaping, and constraint-aware optimization, it introduces an explicit mechanism for balancing engagement and empathy. Although the present work focuses on behavioral health, the underlying paradigm is generalizable to adjacent domains such as education, digital therapeutics, and well-being recommender systems. Future research may extend this foundation through real-world pilots, hybrid human–AI decision architectures, and continual learning under privacy-preserving constraints.

\section{Conclusions}
\label{sec:conclusion}
This paper introduces the Responsible Reinforcement Learning (RRL) framework, a principled approach for aligning adaptive decision-making with emotional, engagement and ethical oriented objectives in behavioral health support systems. Unlike traditional recommendation or prioritization models that optimize short-term engagement, RRL formulates the task as a Constrained Markov decision process (CMDP), explicitly incorporating affective and ethical costs into the optimization objective. Through this structure, the agent is not merely trained to maximize behavioral metrics, but to respect emotional congruence and safety constraints that are essential for long-term trust and well-being.

The proposed architecture further integrates emotional representation learning within the state space, enabling the agent to interpret contextual signals such as mood, tone, and emotional readiness to engage. This design marks a shift from purely utilitarian personalization to empathetic personalization—where user state interpretation becomes a shared space between machine inference and human values. Although our experiments rely on simulated environments, the conceptual foundations offer a robust pathway toward real-world deployment with interpretable, accountable, and ethically grounded adaptive systems.

Future work will extend this framework along several directions. First, hybridizing RRL with foundation models may enable richer contextual reasoning about emotional states and intents. Second, integrating privacy-preserving mechanisms such as federated or split learning will support cross-institutional deployment without compromising confidentiality. Finally, longitudinal field studies in digital therapeutics could quantify how emotionally aligned policies influence engagement, adherence, and recovery trajectories across diverse populations.

\vspace{1em}
\subsubsection*{Broader Impact Statement}
The Responsible Reinforcement Learning framework has the potential to transform how adaptive systems interact with users in sensitive contexts such as mental health, addiction recovery, and chronic disease management. By design, the model prioritizes ethical safety, emotional congruence, and transparency. However, potential risks include overreliance on automated emotional inference, misinterpretation of affective signals, and inadvertent reinforcement of cultural or demographic bias. To mitigate these, deployment should include ongoing human oversight, participatory co-design with clinicians and patients, and periodic fairness auditing. Ultimately, the broader vision is to advance artificial intelligence that not only learns effectively but also cares responsibly—balancing optimization with empathy.

\vspace{1em}
\subsubsection*{Author Contributions}
Conceptualization, methodology, and writing were led by the primary author. The work represents an independent research contribution inspired by challenges in ethical AI and behavioral personalization. No proprietary data or organizational assets were used in this study.

\vspace{1em}
\subsubsection*{Acknowledgments}
The author thanks colleagues and mentors who contributed conceptual insights during informal discussions on reinforcement learning, affective computing, and responsible AI design. This work received no external funding and reflects the author’s independent research perspective.

\bibliography{main}

\begin{thebibliography}{30}
\providecommand{\natexlab}[1]{#1}
\providecommand{\url}[1]{\texttt{#1}}
\expandafter\ifx\csname urlstyle\endcsname\relax
  \providecommand{\doi}[1]{doi: #1}\else
  \providecommand{\doi}{doi: \begingroup \urlstyle{rm}\Url}\fi

\bibitem[Achiam et~al.(2017)Achiam, Held, Tamar, and Abbeel]{achiam2017constrained}
Joshua Achiam, David Held, Aviv Tamar, and Pieter Abbeel.
\newblock Constrained policy optimization.
\newblock In \emph{Proceedings of the 34th International Conference on Machine Learning (ICML ’17)}, volume~70 of \emph{Proceedings of Machine Learning Research}, pp.\  22--31, 2017.
\newblock URL \url{https://proceedings.mlr.press/v70/achiam17a/achiam17a.pdf}.
\newblock First deep-RL method for CMDPs with guarantee on constraint satisfaction.

\bibitem[Bickmore \& Picard(2005)Bickmore and Picard]{bickmore2005relational}
Timothy~W. Bickmore and Rosalind~W. Picard.
\newblock Establishing and maintaining long-term human–computer relationships.
\newblock \emph{ACM Transactions on Computer-Human Interaction (TOCHI)}, 12\penalty0 (2):\penalty0 293--327, 2005.
\newblock \doi{10.1145/1067860.1067867}.

\bibitem[Bickmore et~al.(2010)Bickmore, Schulman, and Yin]{bickmore2010maintaining}
Timothy~W. Bickmore, Daniel Schulman, and Langxuan Yin.
\newblock Maintaining engagement in long-term interventions with relational agents.
\newblock \emph{Applied Artificial Intelligence}, 24\penalty0 (6):\penalty0 648--666, 2010.
\newblock \doi{10.1080/08839514.2010.492259}.

\bibitem[Bond et~al.(2023)Bond, Mulvenna, Potts, O’Neill, Ennis, and Torous]{bond2023digital}
Raymond~R Bond, Maurice~D Mulvenna, Courtney Potts, Siobhan O’Neill, Edel Ennis, and John Torous.
\newblock Digital transformation of mental health services.
\newblock \emph{Npj mental health research}, 2\penalty0 (1):\penalty0 13, 2023.

\bibitem[Chow et~al.(2018)Chow, Nachum, Duenez-Guzman, and Ghavamzadeh]{chow2018lyapunovSafeRL}
Yinlam Chow, Ofir Nachum, Edgar Duenez-Guzman, and Mohammad Ghavamzadeh.
\newblock A lyapunov-based approach to safe reinforcement learning.
\newblock In \emph{Advances in Neural Information Processing Systems (NeurIPS 2018)}, 2018.

\bibitem[Den~Hengst et~al.(2020)Den~Hengst, Grua, el~Hassouni, and Hoogendoorn]{den2020reinforcement}
Floris Den~Hengst, Eoin~Martino Grua, Ali el~Hassouni, and Mark Hoogendoorn.
\newblock Reinforcement learning for personalization: A systematic literature review.
\newblock \emph{Data Science}, 3\penalty0 (2):\penalty0 107--147, 2020.

\bibitem[Keerthana \& Gupta(2025{\natexlab{b}})Keerthana and Gupta]{keerthana2025dense}
Garapati Keerthana and Manik Gupta.
\newblock Dense: Longitudinal progress note generation with temporal modeling of heterogeneous clinical notes across hospital visits.
\newblock \emph{arXiv preprint arXiv:2507.14079}, 2025{\natexlab{b}}.

\bibitem[Garc{\'\i}a \& Fern{\'a}ndez(2015)Garc{\'\i}a and Fern{\'a}ndez]{garcia2015comprehensive}
Javier Garc{\'\i}a and Fernando Fern{\'a}ndez.
\newblock A comprehensive survey on safe reinforcement learning.
\newblock \emph{Journal of Machine Learning Research}, 16\penalty0 (1):\penalty0 1437--1480, 2015.
\newblock URL \url{http://jmlr.org/papers/volume16/garcia15a/garcia15a.pdf}.

\bibitem[{Government of India, Ministry of Electronics \& Information Technology}(2025)]{indiaAI2025governance}
{Government of India, Ministry of Electronics \& Information Technology}.
\newblock \emph{India AI Governance Guidelines: Enabling Safe and Trusted AI Innovation}.
\newblock Press Information Bureau (PIB), Nov 2025.
\newblock URL \url{https://static.pib.gov.in/WriteReadData/specificdocs/documents/2025/nov/doc2025115685601.pdf}.
\newblock Released via PIB, Government of India.

\bibitem[Grabb et~al.(2024)Grabb, Lamparth, and Vasan]{grabb2024risks}
Declan Grabb, Max Lamparth, and Nina Vasan.
\newblock Risks from language models for automated mental healthcare: Ethics and structure for implementation.
\newblock \emph{arXiv preprint arXiv:2406.11852}, 2024.

\bibitem[Hegde \& Jayalath(2025)Hegde and Jayalath]{affective_review_2025}
Karishma Hegde and Hemadri Jayalath.
\newblock Emotions in the loop: A survey of affective computing for emotional support.
\newblock \emph{arXiv preprint arXiv:2505.01542}, 2025.
\newblock URL \url{https://arxiv.org/abs/2505.01542}.

\bibitem[Hornstein et~al.(2023)Hornstein, Zantvoort, Lueken, Funk, and Hilbert]{hornstein2023personalization}
Silvan Hornstein, Kirsten Zantvoort, Ulrike Lueken, Burkhardt Funk, and Kevin Hilbert.
\newblock Personalization strategies in digital mental health interventions: a systematic review and conceptual framework for depressive symptoms.
\newblock \emph{Frontiers in digital health}, 5:\penalty0 1170002, 2023.

\bibitem[Islam et~al.(2025)Islam, Barek, Shahriar, Francia~III, and Ahamed]{islam2025reinforcement}
A.~B. M. Kamrul~Riad Islam, Md.~Abdul Barek, Hossain Shahriar, Guillermo Francia~III, and Sheikh~Iqbal Ahamed.
\newblock Reinforcement learning in medical imaging: Taxonomy, llms, and clinical challenges.
\newblock \emph{Future Internet}, 17\penalty0 (9):\penalty0 396, 2025.
\newblock \doi{10.3390/fi17090396}.

\bibitem[Keerthana \& Gupta(2025{\natexlab{a}})Keerthana and Gupta]{keerthana2025cli}
Garapati Keerthana and Manik Gupta.
\newblock Cli-rag: A retrieval-augmented framework for clinically structured and context aware text generation with llms.
\newblock \emph{arXiv preprint arXiv:2507.06715}, 2025{\natexlab{a}}.

\bibitem[Krasowski et~al.(2023)Krasowski, Thumm, M{\"u}ller, Sch{\"a}fer, Wang, and Althoff]{krasowski2023provably}
Hanna Krasowski, Jakob Thumm, Marlon M{\"u}ller, Lukas Sch{\"a}fer, Xiao Wang, and Matthias Althoff.
\newblock Provably safe reinforcement learning: Conceptual analysis, survey, and benchmarking.
\newblock \emph{Transactions on Machine Learning Research}, 2023.
\newblock Preprint available online.

\bibitem[Li et~al.(2010)Li, Chu, Langford, and Schapire]{li2010contextual}
Lihong Li, Wei Chu, John Langford, and Robert~E Schapire.
\newblock A contextual-bandit approach to personalized news article recommendation.
\newblock In \emph{Proceedings of the 19th international conference on World wide web}, pp.\  661--670, 2010.

\bibitem[Liu et~al.(2024)Liu, Wang, Zhou, Li, Hou, Zhou, Wang, Hoetzlein, and Zhang]{liu2024review}
Ying Liu, Haozhu Wang, Huixue Zhou, Mingchen Li, Yu~Hou, Sicheng Zhou, Fang Wang, Rama Hoetzlein, and Rui Zhang.
\newblock A review of reinforcement learning for natural language processing, and applications in healthcare.
\newblock \emph{Journal of the American Medical Informatics Association}, 31\penalty0 (10):\penalty0 2379--2393, 2024.
\newblock \doi{10.1093/jamia/ocae215}.

\bibitem[Nelekar et~al.(2022)Nelekar, Abdulrahman, Gupta, and Richards]{nelekar2022effectiveness}
Shreeya Nelekar, Amal Abdulrahman, Manik Gupta, and Deborah Richards.
\newblock Effectiveness of embodied conversational agents for managing academic stress at an indian university (aru) during covid-19.
\newblock \emph{British Journal of Educational Technology}, 53\penalty0 (3):\penalty0 491--511, 2022.

\bibitem[Picard(1997)]{picard1997affective}
Rosalind~W. Picard.
\newblock \emph{Affective Computing}.
\newblock MIT Press, Cambridge, MA, 1997.
\newblock ISBN 978-0262011532.

\bibitem[Rodr{\'\i}guez-Ch{\'\i}a et~al.(2022)]{rodriguezchia2022multiobjective}
Javier Rodr{\'\i}guez-Ch{\'\i}a et~al.
\newblock Multi-objective reinforcement learning: A review and new perspective.
\newblock In \emph{Proceedings of the 21st International Conference on Autonomous Agents and Multiagent Systems (AAMAS 2022)}, pp.\  1700--1708, 2022.

\bibitem[Schlicher et~al.(2025)Schlicher, Li, Murthy, Sun, and Schuller]{schlicher2025emotionally}
Michelle Schlicher, Yupei Li, Sunil~M.K. Murthy, Qiyang Sun, and Björn~W. Schuller.
\newblock Emotionally adaptive support: A narrative review of affective computing for mental health.
\newblock \emph{Frontiers in Digital Health}, 7:\penalty0 1657031, 2025.
\newblock \doi{10.3389/fdgth.2025.1657031}.

\bibitem[Keerthana \& Gupta(2025{\natexlab{c}})Keerthana and Gupta]{keerthana2025trimedprompt}
Garapati Keerthana and Manik Gupta.
\newblock Trimedprompt: A unified prompting framework for realistic and layout-conformant clinical progress note synthesis.
\newblock \emph{Journal of Biomedical Informatics}, pp.\  104927, 2025{\natexlab{c}}.

\bibitem[Schuller \& et~al.(2024)Schuller and et~al.]{schuller2024multimodal}
Björn~W. Schuller and et~al.
\newblock Improving access trust in healthcare through multimodal deep learning for affective computing.
\newblock \emph{Human-Centric Intelligent Systems}, 4:\penalty0 511--526, 2024.
\newblock \doi{10.1007/s44230-024-00080-4}.

\bibitem[Stray et~al.(2024)Stray, Halevy, Assar, Hadfield-Menell, Boutilier, Ashar, Bakalar, Beattie, Ekstrand, Leibowicz, et~al.]{stray2024building}
Jonathan Stray, Alon Halevy, Parisa Assar, Dylan Hadfield-Menell, Craig Boutilier, Amar Ashar, Chloe Bakalar, Lex Beattie, Michael Ekstrand, Claire Leibowicz, et~al.
\newblock Building human values into recommender systems: An interdisciplinary synthesis.
\newblock \emph{ACM Transactions on Recommender Systems}, 2\penalty0 (3):\penalty0 1--57, 2024.

\bibitem[Tabassi(2023)]{tabassi2023aiRMF}
Elham Tabassi.
\newblock Artificial intelligence risk management framework (ai rmf 1.0).
\newblock Technical report, National Institute of Standards and Technology (NIST), 2023.
\newblock URL \url{https://doi.org/10.6028/NIST.AI.100-1}.
\newblock Voluntary framework for trustworthy and responsible AI systems.

\bibitem[Valentine et~al.(2023)Valentine, D’Alfonso, and Lederman]{valentine2023recommender}
Lee Valentine, Simon D’Alfonso, and Reeva Lederman.
\newblock Recommender systems for mental health apps: advantages and ethical challenges.
\newblock \emph{AI \& society}, 38\penalty0 (4):\penalty0 1627--1638, 2023.

\bibitem[Wachi et~al.(2024)Wachi, Shen, and Sui]{wachi2024survey}
Akifumi Wachi, Xun Shen, and Yanan Sui.
\newblock A survey of constraint formulations in safe reinforcement learning.
\newblock In \emph{Proceedings of the 33rd International Joint Conference on Artificial Intelligence (IJCAI ’24) – Survey Track}, pp.\  8262--8271, 2024.
\newblock URL \url{https://doi.org/10.24963/ijcai.2024/913}.
\newblock Recent review of constraint formulations in Safe RL emphasising CMDP, shielding etc.

\bibitem[Wang et~al.(2022)Wang, Song, Tao, Liotta, Yang, Li, Gao, Sun, Ge, Zhang, and Zhang]{affective_survey_2022}
Yan Wang, Wei Song, Wei Tao, Antonio Liotta, Dawei Yang, Xinlei Li, Shuyong Gao, Yixuan Sun, Weifeng Ge, Wei Zhang, and Wenqiang Zhang.
\newblock A systematic review on affective computing: Emotion models, databases, and recent advances.
\newblock \emph{arXiv preprint arXiv:2203.06935}, 2022.
\newblock URL \url{https://arxiv.org/abs/2203.06935}.

\bibitem[Yu et~al.(2021)Yu, Liu, Nemati, and Yin]{yu2021reinforcement}
Chao Yu, Jiming Liu, Shamim Nemati, and Guosheng Yin.
\newblock Reinforcement learning in healthcare: A survey.
\newblock \emph{ACM Computing Surveys (CSUR)}, 55\penalty0 (1):\penalty0 1--36, 2021.

\bibitem[Zhao et~al.(2023)Zhao, He, Chen, Wei, and Liu]{zhao2023statewise}
Weiye Zhao, Tairan He, Rui Chen, Tianhao Wei, and Changliu Liu.
\newblock State-wise safe reinforcement learning: A survey.
\newblock In \emph{Proceedings of the 32nd International Joint Conference on Artificial Intelligence (IJCAI ’23) – Survey Track}, pp.\  6814--6822, 2023.
\newblock URL \url{https://doi.org/10.24963/ijcai.2023/763}.
\newblock Survey of state-wise constraints in Safe RL (SCMDPs).

\end{thebibliography}
\bibliographystyle{tmlr}
\end{document}